\documentclass{article}

\usepackage{PRIMEarxiv}

\usepackage[utf8]{inputenc} 
\usepackage[T1]{fontenc}    
\usepackage{hyperref}       
\usepackage{url}            
\usepackage{booktabs}       
\usepackage{amsfonts}       
\usepackage{nicefrac}       
\usepackage{microtype}      
\usepackage{lipsum}
\usepackage{fancyhdr}       
\usepackage{graphicx}       
\graphicspath{{media/}}     

\usepackage{amsmath}
\usepackage{amssymb}
\usepackage{mathtools}
\usepackage{amsthm}

\usepackage{algorithm}
\usepackage{algpseudocode}

\usepackage[capitalize,noabbrev]{cleveref}

\theoremstyle{plain}
\newtheorem{theorem}{Theorem}[section]
\newtheorem{proposition}[theorem]{Proposition}

\theoremstyle{definition}
\newtheorem{definition}[theorem]{Definition}

\theoremstyle{remark}

\title{Shapley Sets: Feature Attribution via Recursive Function Decomposition

}

\author{
  Torty Sivill \\
  University of Bristol \\
  The Alan Turing Institute \\
  vs14980@bristol.ac.uk\\
   \And
  Peter Flach\\
  University of Bristol \\
  The Alan Turing Institute \\
}

\begin{document}
\maketitle

\begin{abstract}
Despite their ubiquitous use, Shapley value feature attributions can be misleading due to feature interaction in both model and data. We propose an alternative attribution approach, Shapley Sets, which awards value to sets of features. Shapley Sets decomposes the underlying model into non-separable variable groups using a recursive function decomposition algorithm with log linear complexity in the number of variables. Shapley Sets attributes to each non-separable variable group their combined value for a particular prediction. We show that Shapley Sets is equivalent to the Shapley value over the transformed feature set and thus benefits from the same axioms of fairness. Shapley Sets is value function agnostic and we show theoretically and experimentally how Shapley Sets avoids pitfalls associated with Shapley value based alternatives and are particularly advantageous for data types with complex dependency structure.
\end{abstract}


\section{The Shapley Value and Non-separable Functions}

In co-operative game theory, one central question is that of fair division: if players form a coalition to achieve a common goal, how should they split the profits? Let $N$ be the set $\{1, 2, ... n \}$ of players and $2^N$ all coalitions of players. A function $v : 2^{N} \rightarrow \mathbb{R}$ is the $n$-person game in characteristic form, such that $v(S), S \subseteq N$ defines the worth of coalition $S$ where $v(\varnothing) = 0$. A solution concept is a mapping assigning a vector $ \mathbf{x} \in \mathbb{R}^{n}$ to the game $v$. The Shapley value \cite{shapley1997value} is the most widely known solution concept which uniquely satisfies certain axioms of fairness: efficiency, dummy, symmetry and additivity. Please see \cite{shapley1997value} for definitions.

\begin{definition}[Shapley Value]
For the game $v$ the Shapley value of player $i \in N$ is given as
\begin{equation}
\phi_{i}(v) = \Sigma_{S \subseteq N \backslash \{i\}} \frac{|S|! (n-|S|-1)!}{n!} [v(S \cup \{i\}) - v(S)]
\end{equation}
\end{definition}

Under efficiency, the Shapley value decomposes the value of the grand coalition $v(N) - v(\varnothing)$ to attribute worth to each individual player. The Shapley value is a fully separable function (Definition \ref{def_additive}) such that $v(N) - v(\varnothing) = \sum_{i}^{n} \phi_{i}(v)$.

\begin{definition} \label{def_additive}[Additively Separable Function]
A function $f : \mathbb{R}^{n} \rightarrow \mathbb{R}$ with variable set $\mathbf{X} =\{X_{1},...,X_{n}\}$ is separable if it has the following form
$f(\mathbf{X}) = \sum_{i =1}^{k} f_{i}(\mathbf{X}_{i})  $
$1 < k \leq n$,

where  $\mathbf{X}_{1}, \mathbf{X}_{2},...,\mathbf{X}_{k}$ are $k$ non-overlapped sub-vectors of $\mathbf{X}$.
\end{definition}

Specifically, the function $f$ is also called fully additively separable if $k = n$, while it is regarded as fully non-separable if $k = 1$. While there are other forms of separability, In this paper we use the term separable to refer to additive separability.

The set function $v$ may not be fully separable. Within coalitional games this is due to the interaction between players. Consider the following example for the game $v$ with player set $ N = \{1,2,3\}$ and $v(1) = 1,v(2) = 0,v(3) = 0, v(1,2) = 1, v(1,3) = 1, v(2,3) = 2, v(1,2,3) = 3$. Clearly the game is not fully separable as $v(1) + v(2) + v(3) \not = v(1,2,3)$. The non-separable interaction effects within coalitional games are dealt with by solution concepts which map partially separable into fully separable functions, allowing an individual attribution of worth to each player. The Shapley value provides an attribution where each player receives an average of their marginal contribution to all coalitions. 

\subsection{Interaction Effects For Feature Attribution}

When applying the Shapley value to feature attribution, there are three functions to consider: The model to be explained $f: \mathbb{R}^{n} \rightarrow \mathbb{R}$ which operates on a variable set $\mathbf{X} = \{X_{1},...,X_{n}\}$. The set function $v$, which takes as input a set of features $\mathbf{X}_{S} \subseteq \mathbf{X}$ and obtains $f$'s prediction on this coalition of features. The Shapley value $\phi(v)$ which maps the set function $v$ into a fully  separable function. Given a particular prediction to attribute, $f(\mathbf{x})$ where $\mathbf{x}=\{x_{1},...,x_{n}\}$, the value function $v(\mathbf{x},\mathbf{X}_{S})$ specifies how the subset of features, $\mathbf{X}_{\bar{S} } = \mathbf{X} \backslash \mathbf{X}_{S}$, should be removed from $\mathbf{x}$. 

In coalitional game theory, the Shapley value attributes the difference in value between the grand coalition and the empty set of players. For feature attribution, the Shapley value attributes the change in prediction between instance and baseline. Therefore, the value of the empty coalition, $v(\mathbf{x},\mathbf{X}_{\varnothing})$ is not guaranteed to be zero but some uninformative baseline prediction and thus the value function must account for the non-zero baseline: $v(\mathbf{x},\mathbf{X}_{S}) = f(\mathbf{X}_{\bar{S}},\mathbf{X}_{S}=\mathbf{x}_{S}) - f(\mathbf{X}_{\varnothing})$. Similarly to coalitional games, the Shapley value fairly allocates interaction effects to each feature. Feature interaction may occur in the data $f(X_{1},X_{2},X_{3}) = X_{1} + X_{2} + X_{3}$ where $X_{2} = \alpha X_{3}$ and/or in the model $f(X_{1},X_{2},X_{3}) = X_{1} + 2 X_{2}X_{3}$. The choice of value function, which acts as the interface between the Shapley value and the function $f$ determines the kind of interactive effects the Shapley value must allocate between features. 
 
\subsection{When Interaction Occurs in the Data}
\label{interaction_in_data}

While the following ideas have previously been discussed \cite{kumar2020problems,aas2021explaining,heskes2020causal}, we re-frame them here within the context of separability which allows us to motivate our proposed attribution method, Shapley Sets. 

\textbf{Example 1:} Given the binary variable set  $\mathbf{X} = \{X_{1},X_{2},X_{3}\}$ and function $f(\mathbf{X}) = X_{1} + X_{3}$ where $X_{2}$ is the causal ancestor of $X_{3}$ such that $X_{3} = X_{2}$.
It is clear that $X_{2}$ has no impact on $f(\mathbf{X})$ from the perspective of the model. However, from the perspective of the data distribution, $X_{3}$ is dependent on $X_{2}$. Changing $X_{2}$ will result in a change in $X_{3}$ therefore, changing $X_{2}$ to a value non-consistent with $X_{3}$ does not make sense. Whether to consider $X_{2}$ as a separate player in the game and attribute value despite it having no direct influence on the model output is an open debate in the literature.

\textbf{Off-manifold Value Functions} There are those who argue that features with no impact on the model should receive no attribution \cite{merrick2020explanation,janzing2020feature}. These methods break all statistical relationships between the inputs to the model by using a value function which calculates the impact of each feature on the model independently of its impact on the distribution of other features. This approach was formalised as $v_{marg}$ by \cite {janzing2020feature}
\begin{equation}
\label{eq_vmarg}
v_{marg}(\mathbf{x},\mathbf{X}_{S}) = 
f(\mathbf{X}_{S}=\mathbf{x}_{S},\mathbb{E}[\mathbf{X}_{\bar{S}}]) - f(\mathbb{E}[\mathbf{X}])
\end{equation} The expectation is usually taken over the input distribution $\mathbf{X}_{input}$. However, if this is replaced by an arbitrary distribution, $v_{marg}$ was generalised to $v_{bs}$ by \cite{sundararajan2020many} which uses an arbitrary baseline sample $\mathbf{z}$, 
\begin{equation}
\label{equation_vbs}
    v_{bs}(\mathbf{x},\mathbf{z},\mathbf{X}_{S}) = f(\mathbf{X}_{S} = \mathbf{x}_{S}, \mathbf{X}_{\bar{S}} = \mathbf{z}_{\bar{S}}) - f(\mathbf{X}=\mathbf{z}).
\end{equation}

There are those who argue that attributions independent of the statistical interactions in the data are inherently misleading \cite{hooker2021unrestricted,heskes2020causal}. Firstly from a causal perspective, if we consider Example 1, the Shapley value via $v_{marg}$ would assign zero importance to $X_{2}$. An attribution ignoring that $X_{2}$ is directly responsible for $X_{3}$ is misleading, especially if the attribution is used to recommend changes. Furthermore, $v_{marg}$ evaluates the model on out of distribution samples. If we break the causal relationship between $X_{2}$ and $X_{3}$, and using their independent expected values $\mathbb{E}[X_{2}]=\mathbb{E}[X_{3}]=0$ in $v_{marg}$, the model is evaluated on samples $(x_{1},1,0)$ which is a complete misrepresentation of the truth. 

\textbf{On-manifold Value Functions} To combat this problem, on-manifold samples can be calculated by the use of the conditional value function, first introduced by \cite{vstrumbelj2014explaining} which does consider statistically related features as separate players in the game, allowing the distribution of out of coalition features to be impacted by the feature in question
\begin{equation} \label{eq_vcond} v_{cond}(\mathbf{x},\mathbf{X}_{S}) = \mathbb{E}[f(\mathbf{X}_{S}=\mathbf{x}_{S},\mathbf{X}_{\bar{S}})|\mathbf{X}_{S}=\mathbf{x}_{S}] - \mathbb{E}[f(\mathbf{X})].
\end{equation}
$v_{cond}$ is often taken as the observational conditional probability whereby the expected conditional is calculated over $\mathbf{X}_{input}$. This generates on-manifold data samples which address the problems discussed above. Furthermore, features which have no direct impact on the model but an indirect impact through other features are assigned a non zero importance, more accurately reflecting reality. However, the two significant issues with $v_{cond}$ are its computational complexity: requiring the evaluation of the model on $2^{N}$ multivariate conditional distributions and the undesirable impact of considering all features as players combined with the efficiency which we explicate below.

In assigning non-marginal features a non zero importance, $v_{cond}$ can give misleading explanations which indicate features to change despite having zero impact on the outcome. This weakness of $v_{cond}$ has been formalised as a ``violation of sensitivity'' \cite{janzing2020feature}: \textit{When the relevance of $\phi_{i}$ is defined by $v_{cond}, \phi_{i} \not = 0$ does not imply that $f$ depends on $X_{i}$}

The failure of sensitivity exhibited by $v_{cond}$ leads to further issues with the generated attributions. Consider Example 1 again, where $X_{3},X_{2},X_{1}$ are binary variables and $X_{3} = X_{2}$.
Given the input $\mathbf{x} = (x_{1},x_{2},x_{3}) =(1,1,1)$ and $f(x_{1},x_{2},x_{3})=2$ Under $v_{cond}$, the Shapley attributions for $X_{2}$ and $X_{3}$ would both be greater than the attribution for $X_{1}$. Clearly, the attribution of $X_{2}$ violates sensitivity. Now, consider the an alternative function which is just trained on two features $X_{1}, X_{3}$. As $X_{3}=X_{2}$ $f_{2}(X_{1},X_{3}) = f(X_{1},X_{2},X_{3})$. However, now the Shapley values for $X_{1}$ and $X_{3}$ are equal. The relative apparent importances of $X_{1}$ and $X_{3}$ depend on whether $X_{2}$ is considered to be a third feature, even though the two functions are effectively the same. 

\cite{frye2020asymmetric} propose a solution to the failure of sensitivity exhibited by $v_{cond}$ following the intuition: If $X_{i}$ is known to be the deterministic causal ancestor of $X_{j}$, one might want to attribute all effect to $X_{i}$ and none to $X_{j}$.
In contrast, \cite{heskes2020causal} argue that the only way to remove the problems arising from the failure of sensitivity is to replace the observational $v_{cond}$ with the interventional conditional distribution. 
 However, both the asymmetric and interventional attribiutions above require the specification of the causal structure of the phenomenon being modelled. It has been argued \cite{kumar2020problems} that this requirement is a significant limiting factor in the adoption of either approach.  
In this paper, we propose an attribution approach which can be used with on and off-manifold value functions. Under $v_{cond}$, our method generates on-manifold attributions which avoid the failure of sensitivity without requiring any knowledge of the causal structure of the underlying data distribution. 

\subsection{When Interaction Occurs in the Model}

While off-manifold value functions ignore interaction in the data, both on and off-manifold value functions recognize interaction in the model. It has been recognised, however that the Shapley value, in the presence of feature interaction in the model, generates misleading attributions \cite{kumar2021shapley}. 

\textbf{Example 2:} Consider the function $f(X_{1},X_{2},X_{3}) = X_{1} + 2 X_{2}X_{3}$ and assume that each of the three features are statistically independent, i.e. all interaction between features is defined entirely by the model. Furthermore, it is given that $\mathbb{E}[X_{1}] = \mathbb{E}[X_{2}] = \mathbb{E}[X_{3}] = 0$ and that our sample to be explained $\mathbf{x} = (1,1,1)$. The Shapley value under both on and off-manifold value functions give equal attributions to each feature. While this attribution makes sense from the perspective of how much each feature contributed to the change in prediction, it does not reflect the true behaviour of the model where changing the value of $X_{2}$ or $X_{3}$ would have double the impact on the model as changing $X_{1}$. In this paper, we propose a solution concept which would group $X_{2},X_{3}$ and unlike the Shapley value, award attribution together, resulting in attributions more faithful to the underlying model $f$ when used with on or off-manifold value functions. 

\section{Shapley Sets of Non-Separable Variable Groups}

The problems with Shapley value attributions discussed above occur as it assigns individual value to variables belonging to Non-Separable Variable Groups (NSVGs) in regards to the underlying partially separable function $f$ (Definition \ref{def_additive}). Non-separable groups are used to describe the formed variable groups $\{\mathbf{X}_{1},...,\mathbf{X}_{k}\}$ after a complete (or ideal) decomposition of $f$. An NSVG can also be defined as the minimal set of all interacted variables given the function $f$ which we explicate in Definition \ref{def_non-sep-variable}. 

\begin{definition}[Non-Separable Variable Group (NSVG)] \label{def_non-sep-variable}

Let $\mathbf{X} = \{X_{1},X_{2},...X_{n}\}$ be the set of decision variables and $f$ be a partially separable function $f : \mathbb{R}^{n} \rightarrow \mathbb{R}$ satisfying Definition \ref{def_additive}. If there exists any two candidate decision vectors $\mathbf{x} = \{x_{1},...,x_{n}\}$ and $\mathbf{x}' = \{x'_{1},...,x'_{n}\}$, sampled from the domain of $\mathbf{X}$, such that the following property holds for any two mutually exclusive subsets $\mathbf{X}_{i}, \mathbf{X}_{j} \subset \mathbf{X}$, $\mathbf{X}_{i} \cap \mathbf{X}_{j} = \varnothing$,


\begin{equation}
\label{eq_interaction}
f(\mathbf{x})_{\mathbf{X}_{i} \cup \mathbf{X}_{j}} - f(\mathbf{x})_{\mathbf{X}_{j}}
\not = f(\mathbf{x})_{\mathbf{X}_{i}} - f(\mathbf{x})_{\mathbf{X}_{\varnothing}},
\end{equation}

then the sets $\mathbf{X}_{i},\mathbf{X}_{j}$ are said to interact. Here, $f(\mathbf{x})_{\mathbf{X}_{S}} = f(\mathbf{X}_{S} = \mathbf{x}_{S}, \mathbf{X}_{\bar{S}} = \mathbf{x'}_{\bar{S}})$ and $\mathbf{X}_{S} \cup \mathbf{X}_{\bar{S}} = \mathbf{X}$. As an NSVG refers to the minimal set of interacted variables, if $|\mathbf{X}_{i}|$ and $|\mathbf{X}_{j}|$ is minimized such that Equation \ref{eq_interaction} still holds, then $\mathbf{X}_{i} \cup \mathbf{X}_{j}$ is a NSVG. (For proof, see \cite{sun2017recursive}).

\end{definition}
Translating Definition \ref{def_non-sep-variable} for feature attribution, given that $f(\mathbf{x})_{\mathbf{X}_{S}}$ is a function over the domain of all the possible subsets of $\mathbf{X}_{S} \subseteq \mathbf{X}$ we can rewrite Equation \ref{eq_interaction} in terms of  $v(\mathbf{x},\mathbf{X}_{S})$, where $v$ could represent any of the value functions from the previous section but in this paper we restrict $v \in \{v_{cond},v_{bs}\}$. By setting $\mathbf{X}_{i}=\{X_{i}\}$ and $\mathbf{X}_{j}= \mathbf{X}_{S}$, for $v_{bs}$, given that $|\mathbf{X}_{S}|$ is minimised, if there exist any candidate vectors $\mathbf{x},\mathbf{x'}$ such that 
\begin{equation}
\label{eq_interaction_bs}
v_{bs}(\mathbf{x},\mathbf{x'},\{X_{i}\} \cup \mathbf{X}_{S}) - v_{bs}(\mathbf{x},\mathbf{x'},\mathbf{X}_{S}) \not = v_{bs}(\mathbf{x},\mathbf{x'},\{X_{i}\} )
\end{equation} 
then $\{X_{i}\} \cup \mathbf{X}_{S}$ is a NSVG.

For $v_{cond}$, given that $|\mathbf{X}_{S}|$ is minimised, if there exists any candidate vector $\mathbf{x}$ such that
\begin{equation}
\label{eq_interaction_cond}
v_{cond}(\mathbf{x},\{X_{i}\} \cup \mathbf{X}_{S}) - v_{cond}(\mathbf{x},\mathbf{X}_{S}) \not = v_{cond}(\mathbf{x},\{X_{i}\})\end{equation}
then $\{X_{i}\} \cup \mathbf{X}_{S}$ is a NSVG. 

Given the partially separable function from Example 2, under $v_{bs}$, $\{X_{2},X_{3}\}$ is a NSVG as $v_{bs}(\mathbf{x},\mathbf{x'},\{X_{3},X_{2}\}) - v_{bs}(\mathbf{x},\mathbf{x'},\{X_{2}\}) \not = v_{bs}(\mathbf{x},\mathbf{x'},\{X_{3}\})$ for settings $\mathbf{x}=(1,1,1)$ and $\mathbf{x'}=(0,0,0)$.

Given the partially separable function from Example 1, under $v_{cond}$, the set $\{X_{2},X_{3}\}$ is a NSVG as $v_{cond}(\mathbf{x},\{X_{3},X_{2}\}) - v_{cond}(\mathbf{x},\{X_{2}\})
\not = v_{cond}(\mathbf{x},X_{3})$ for setting $\mathbf{x}=(1,1,1)$.

In this paper, we propose an alternative attribution method which, unlike the Shapley value, does not separate NSVGs to assign attribution. We work under the intuition that any interacting feature whether that be in the model or in the data should not be considered as separate players in the coalitional game but should be awarded value together. In both the examples above, $X_{2}$ and $X_{3}$ would receive joint attribution under our proposed method. 

Given the partially separable function $f$ satisfying Definition \ref{def_additive}, variable set $\mathbf{X} = \{X_{1},X_{2},...,X_{n}\}$, and a specified value function $v_{cond,marg}(\mathbf{x},\mathbf{X}_{S})$, our proposed solution concept $\varphi$, which we term Shapley Sets (SS), finds the optimal decomposition of $f$ into the set of $ m > 1$ NSVGs $\{\mathbf{X}_{1},...,\mathbf{X}_{m}\}$. The resulting variable grouping $\{\mathbf{X}_{1},...,\mathbf{X}_{m}\}$ satisfies Definition \ref{def_additive} and each variable group is composed solely of variables $\mathbf{X}_{i}$ which satisfy definition \ref{def_non-sep-variable}. From Definition \ref{def_additive}, $f(\mathbf{x}) = \sum _{i=1}^{m} v(\mathbf{x},\mathbf{X}_{i})$. Given a prediction to be attributed, $f(\mathbf{x})$,  our proposed attribution, $\varphi$, therefore returns the attribution for each variable group $\mathbf{X}_{i}, \forall i \in m$ given as:
\begin{equation}
\label{eq_shapley_sets}
\varphi_{\mathbf{X}_{i}} = v(\mathbf{x},\mathbf{X}_{i})
\end{equation}

\begin{proposition}
\label{lemma_shap_equivalence}
If we model each NSVG, $\mathbf{X}_{i} \in \{\mathbf{X}_{i},...,\mathbf{X}_{m}\}$ as a super-feature $Z_{i}$ such that $\mathbf{Z} = \{Z_{i},...,Z_{m}\}$, $z_{i} = \mathbf{x}_{i}$ and $\mathbf{z} = \{z_{1},...,z_{n}\}$
The Shapley value of each super feature 
 $\phi_{Z_{i}}(v,\mathbf{z})$ is equivalent to $v(\mathbf{z},Z_{i})$

\end{proposition}

\begin{proof}
$$
\phi_{Z_{i}}(v,\mathbf{z}) = \sum_{\mathbf{Z}_{S} \subseteq \mathbf{Z} \backslash \{Z_{i}\}} \alpha [v(\mathbf{z},\{Z_{i}\} \cup \mathbf{Z}_{S})  - v(\mathbf{z},\mathbf{Z}_{S})]
$$
where $\alpha = \frac{|\mathbf{Z}_{S}|! (|\mathbf{Z}|-|\mathbf{Z_{S}}|-1)!}{|\mathbf{Z}|!} $

Given that each $\mathbf{Z}_{i},\mathbf{Z}_{j} \subseteq \mathbf{Z}$ is a NSVG, from Definition \ref{def_non-sep-variable} we know that $v(\mathbf{z},\mathbf{Z}_{i} \cup \mathbf{Z}_{j}) -  v(\mathbf{z},\mathbf{Z}_{j})  = v(\mathbf{z},\mathbf{Z}_{i})$ for any $\mathbf{Z}_{i},\mathbf{Z}_{j} \subseteq{\mathbf{Z}}$. Therefore,  $v(\mathbf{z},\{Z_{i}\} \cup \mathbf{Z}_{S}) - v(\mathbf{z},\mathbf{Z}_{S}) = v(\mathbf{z},\{Z_{i}\})$

It follows that given $\sum_{\mathbf{Z}_{S} \subseteq \mathbf{Z} \backslash \{Z_{i}\}} \frac{|\mathbf{Z}_{S}|! (|\mathbf{Z}|-|\mathbf{Z_{S}}|-1)!}{|\mathbf{Z}|!} = 1 $
$$
\phi_{Z_{i}}(v,\mathbf{z}) = \sum_{\mathbf{Z}_{S} \subseteq \mathbf{Z} \backslash \{Z_{i}\}} \alpha v(\mathbf{z},\{Z_{i}\})   = v(\mathbf{z},\{Z_{i}\}) = v(\mathbf{x},\mathbf{X}_{i})
$$

\end{proof}

Proposition \ref{lemma_shap_equivalence} shows how the attribution given by Shapley Sets (SS) to variable $X_{i}$, $\varphi_{\mathbf{X}_{i}}(v,\mathbf{x})$, is equivalent to the Shapley value when played over the feature set $\mathbf{Z}$ containing the set of NSVGs $\{Z_{1},...,Z_{m}\} = \{\mathbf{X}_{1},...,\mathbf{X}_{m}\}$ for a given $v \in \{v_{cond},v_{bs}\}$. SS therefore satisfies the same axioms of fairness as the Shapley value: efficiency, dummy, additivity and symmetry when played over this feature set. However, we have discussed how, despite its axioms, the Shapley value can generate misleading attributions in the presence of feature interaction. In Section \ref{section_motiv_ss} we therefore give practical advantages of the SS over the Shapley value. First however, we provide a method for finding the optimal decomposition of $f$ into its NSVGs. 

\section{Computing Shapley Sets}





Determining the NSVGs of a function $f$ could be achieved manually by partitioning the variable set and determining interaction over every possible candidate vector. However, this would be computationally intractable. Instead, there exists a large body of literature surrounding function decomposition in global optimization problems. Of this work, automatic decomposition methods identify NSVGs. We therefore propose a method for calculating SS which is based on the Recursive Decomposition Grouping algorithm (RDG) as introduced in \cite{sun2017recursive}. 

To identify whether two sets of variables $\mathbf{X}_{i}$ and $\mathbf{X}_{j}$ interact, RDG uses a fitness measure, based on Definition \ref{def_non-sep-variable} with candidate vectors, $\mathbf{x},\mathbf{x}'$ as the lower and upper bounds of the domain of $\mathbf{X}$. If the difference between the left and right hand side of Equation \ref{eq_interaction} meets some threshold $\epsilon = \alpha \min \{|f(\mathbf{x}_{1})|,...,|f(\mathbf{x}_{k})|\}$ where $\mathbf{x}_{k}$ is a randomly selected candidate vector, then $\mathbf{X}_{i}$ and $\mathbf{X}_{j}$ are deemed by RDG to interact. To adapt RDG for $v_{cond}$ and $v_{bs}$, we propose an alternative fitness measure, Definition \ref{def_ss_fitness}, with candidate vectors $\mathbf{x},\mathbf{x'}$ randomly sampled from $\mathbf{X}_{input}$ which can identify NSVGs in the function and/or in the model.




\begin{definition}[Shapley Sets Fitness Measure]
\label{def_ss_fitness}
Given two sets of variables $\mathbf{X}_{i},\mathbf{X_{j}}$ and a specified value function, $ v \in \{v_{bs},v_{cond}\}$,
if $|v_{cond}(\mathbf{x},\mathbf{X}_{i} \cup \mathbf{X}_{j}) - v_{cond}(\mathbf{x},\mathbf{X}_{j}) -  v_{cond}(\mathbf{x},\mathbf{X}_{i})| > \epsilon$
then there is interaction between $\mathbf{X}_{i}$ and $\mathbf{X}_{j}$.  Or,
if $|v_{bs}(\mathbf{x},\mathbf{x'},\mathbf{X}_{i} \cup \mathbf{X}_{j}) - v_{bs}(\mathbf{x},\mathbf{x'},\mathbf{X}_{j}) -  v_{bs}(\mathbf{x},\mathbf{x'},\mathbf{X}_{i})| > \epsilon$
then there is interaction between $\mathbf{X}_{i}$ and $\mathbf{X}_{j}$. 
\end{definition}

We substitute the SS fitness measure into the RDG algorithm which identifies NSVGs by recursively identifying individual variable sets $\mathbf{X}_{j}$ with which a given variable $X_{i}$ interacts with. If $X_{i}$ and a single variable $X_{j}$ are said to interact they are placed into the same NSVG, $\mathbf{X}_{1}$. At which point conditional interaction between $\mathbf{X}_{1}$ and the remaining variables is identified. The algorithm iterates over every variable $X_{i} \in \mathbf{X}$ and returns the set of NSVGs. To compute the SS attributions for a given prediction $f(\mathbf{x})$ we compute $v(\mathbf{x},\mathbf{X}_{i})$ for each NSVG, $\mathbf{X}_{i}$. Our full algorithm is shown in Algorithm 2. The runtime of SS is $O(n log n)$ as proven in \cite{sun2017recursive}.


\begin{algorithm}[h]
\caption{ValueInteract($\mathbf{X}_{1},\mathbf{X}_{2}$)}\label{alg:cap}
\begin{algorithmic}
\Require $v \in \{v_{bs},v_{cond}\}, \mathbf{X}_{input}, \epsilon$
\If{$v_{bs}$}
    \State Sample $\mathbf{x},\mathbf{x}'$ from input distribution $\mathbf{X}_{input}$
    \State $\sigma_{1} = v(\mathbf{x},\mathbf{x'},\mathbf{X}_{1} \cup \mathbf{X}_{2}) - v(\mathbf{x},\mathbf{x'} \mathbf{X}_{2})$
    \State $\sigma_{2} = v(\mathbf{x},\mathbf{x'},\mathbf{X}_{1})$
\EndIf
\If{$v_{cond}$}
    \State Sample $\mathbf{x}$ from input distribution $\mathbf{X}_{input}$
    \State $\sigma_{1} = v(\mathbf{x},\mathbf{X}_{1} \cup \mathbf{X}_{2}) - v(\mathbf{x}, \mathbf{X}_{2})$
    \State $\sigma_{2} = v(\mathbf{x},\mathbf{X}_{1})$
\EndIf
\If{$\vert \sigma_{1} - \sigma_{2} \vert > \epsilon$}
    \If{$\mathbf{X}_{2}$ contains one variable}
        \State $\mathbf{X}_{1} = \mathbf{X}_{1} \cup \mathbf{X}_{2}$
    \Else
        \State Split $\mathbf{X}_{2}$ into two equal groups $G_{1},G_{2}$
        \State $\mathbf{X}^{1}_{1}$ = ValueInteract($\mathbf{X}_{1},G_{1})$
        \State $\mathbf{X}^{2}_{1}$ = ValueInteract($\mathbf{\mathbf{X}}_{1},G_{2})$
        \State $\mathbf{X}_{1} = \mathbf{X}^{1}_{1} \cup \mathbf{X}^{2}_{1}$
    \EndIf
\EndIf
\State Return $\mathbf{X}_{1}$     
\end{algorithmic}
\end{algorithm}

\begin{algorithm}[t]
\label{alg_ss}
\caption{Shapley Sets (Adapted from RDG \cite{sun2017recursive})}
\begin{algorithmic}
\Require $v \in \{v_{cond},v_{bs}\},\epsilon$, $\mathbf{x}_{inp}$, $\mathbf{x}_{ref}$ (if $v = v_{bs}$)
\State Intialise $seps$ and $nonseps$ as empty groups 
\State Assign the first variable in $\mathbf{X}$ to $\mathbf{X}_{1}$
\State Assign the rest of the variables in $\mathbf{X}$ to $\mathbf{X}_{2}$
\While{$\mathbf{X}_{2}$ is not empty} 
    \State $\mathbf{X'}_{1} \gets ValueInteract(\mathbf{X}_{1},\mathbf{X}_{2})$
    \If{$\mathbf{X'}_{1}$ is the same as $\mathbf{X}_{1}$}
        \If{$\mathbf{X}_{1}$ contains one variable}
            \State $seps \gets X_{1}$
        \Else
            \State $nonseps \gets \mathbf{X}_{1}$
        \EndIf
        \State Empty $\mathbf{X}_{1}$ and $\mathbf{X}'_{1}$
        \State Assign the first variable of $\mathbf{X}_{2}$ to $\mathbf{X}_{1}$
        \State Delete the first variable of $\mathbf{X}_{2}$
    \Else
        \State $\mathbf{X}_{1} = \mathbf{X'}_{1}$
        \State Delete the variables of $\mathbf{X}_{1}$ from $\mathbf{X}_{2}$   
    \EndIf
\EndWhile
\State For each set of variables $\mathbf{X_{i}} \in seps \cup nonseps$ return $v_{cond}(\mathbf{x}_{inp},\mathbf{X}_{i})$ or $v_{bs}(\mathbf{x}_{inp}, \mathbf{x}_{ref}, \mathbf{X}_{i})$  
\end{algorithmic}
\end{algorithm}

\section{Motivating Shapley Sets}
\label{section_motiv_ss}
The selection of the value function $v$ determines the variable grouping generated. Used with $v_{bs}$, as interacting features are placed in the same NSVG, the attributions resulting from SS will be more faithful to the underlying model. The SS attribution for $v_{bs}$ in Example 2 would be $\varphi_{X_{1}} = 1$ and $\varphi_{X_{2},X_{3}} = 2$.

Used with $v_{cond}$, as interacting features are placed in the same NSVG, the attributions resulting from SS do not suffer from the violation of sensitivity as described in Section \ref{interaction_in_data}. Consider again Example 1, as $X_{2},X_{3}$ now belong to a NSVG, the SS attributions for  $X_{1},X_{3}$ are now equal across both $f$ and $f_{2}$  therefore robust to whether non-directly impacting features are included in the model. SS offer a further advantage when used to compare the attributions under on and off-manifold examples. Consider again Example 2 yet now with $X_{1} = \alpha X_{2}$. The SS attribution via $v_{marg}$ would be $\varphi_{X_{1}} = 1$ and $\varphi_{X_{2},X_{3}} = 2$. However, if SS was calculated via $v_{cond}$ $\varphi_{\{X_{1},X_{2},X_{3}\}} = 3 - \mathbb{E}[f(\mathbf{X})]$ indicating that $f$ is non-separable and all the features interact. 

The comparison between on and off-manifold SS therefore indicate \textit{where} the feature interaction takes place. We have thus far provided an alternative attribution method to the Shapley Value, SS which can be computed in $O(n log n)$ time with $n$ being the number of features. SS can be adapted for arbitrary value functions and offers several advantages over Shapley value based attributions when used with on and off-manifold value functions. In Section \ref{sec_experiments} we empirically validate the theoretical claims made above but first we discuss related work. 

\section{Related Work}
As Shapley value based feature attribution has a rich literature, we differentiate SS from three approaches which are closest in essence to ours. SS enforce a coalition structure on the Shapley value such that players cannot be considered in isolation from their coalitions. \textbf{The Owen value} is a solution concept for games with an existing coalition structure \cite{owen1977values}. The Owen value is the result of a two–step procedure: first, the coalitions play a quotient game among themselves, and each coalition receives a payoff which, in turn, is shared among its individual players in an internal game. Both payoffs are given by applying the Shapley value. This approach is not equivalent to SS, who assume no prior coalitional structure, and instead finds the optimum coalition structure which is the decomposition of $v$ into its NSVGs. 

\textbf{Shapley Residuals} \cite{kumar2021shapley} capture the level to which a value function is inessential to a coalition of features. They show for $v_{cond},v_{marg}$, that if the game function can be decomposed into $ v(\mathbf{x},\mathbf{X}_{S}) = v(\mathbf{X}_{T}) + v(\mathbf{X}_{\bar{T}})$ for $\mathbf{X}_{T} \subset \mathbf{X}_{S}$ then the value function $v$ is inessential with respect to the coalition $\mathbf{X}_{T}$. In this way we can view Shapley residuals, $r_{S} \neq 0$ as an indication that a coalition is an non-separable variable group. However, the Shapley residuals are built on complex Hodge decomposition \cite{stern2019hodge}, are difficult to understand and do not offer a better way of attributing to features. In contrast, SS is built on the idea of additive separability, easier to understand, less computationally expensive and propose a solution to the issues with the Shapley value which are analogous to those Shapley residuals were designed to identify. 

\textbf{Grouped Shapley Values} Determining the Shapley value of grouped variables has been previously suggested in \cite{jullum2021groupshapley,NIPS2017_7062} which identify interaction in the data (based on measures of correlation) to then partition the features into groups, after which the Shapley value is then calculated. Shapley Sets is distinct from the above approaches in the following ways. Firstly, Shapley Sets is capable of uncovering interaction in the model as well as in the data. 
Secondly, Shapley Sets is designed to find the optimal grouping of the features such that the Shapley value theoretically reduces to the simple computation in Equation \ref{eq_shapley_sets}. Therefore, the grouping under Shapley Sets requires linear time to compute (given the prior decomposition of the variable set under log linear time), whereas the grouping proposed under grouped shapley values \cite{jullum2021groupshapley,NIPS2017_7062} still requires exponential computation (to compute exactly although this can be approximated). Shapley Sets, to our knowledge, is the first contribution to the feature attribution literature which automatically decomposes a function into the optimal variable set by which to award attribution.

\section{Experimental Motivation of Shapley Sets}
\label{sec_experiments}

We begin with two synthetic experiments. The first of these motivates the use of SS in the presence of interaction in the model. The second motivates the use of the SS in the presence of interaction in the data. We then compare SS to existing Shapley value (SV) based attribution methods on three benchmark datasets. We first however, outline how the value functions $v_{bs},v_{cond}$ are computed for our experiments. 

As discussed above, $v_{bs}$ takes as input arbitrary reference vectors. For our experiments we select $v_{marg}$ such that $v_{bs} = v_{marg}$ (Equation \ref{eq_vmarg}). The expectation is taken over the empirical input distribution $\mathbf{X}_{input}$
For the calculation of $v_{cond}$ (Equation \ref{eq_vcond}), as the true conditional probabilities for the underlying data distribution are unknown we approximate $p(\mathbf{X}_{\bar{S}}| \mathbf{X}_{S} = \mathbf{x}_{s)}$ using the underlying data distribution. Approximating conditional distributions can be achieved by directly sampling from the empirical data distribution. However, as noted in \cite{aas2021explaining}, the this method of approximating $p(\mathbf{X}_{\bar{S}}| \mathbf{X}_{S}) = \mathbf{x}_{s}$ suffers when $|\mathbf{X}_{S}| > 2$, due to sparsity in the underlying empirical distribution. We therefore adopt the approach of \cite{aas2021explaining}, where under the assumption that each $\mathbf{x} \in \mathbf{X}$ is sampled from a multivariate Gaussian with mean vector $\mathbf{\mu}$ and covariance matrix $\mathbf{\Sigma}$, the conditional distribution $ p(\mathbf{X}_{\bar{S}}| \mathbf{X}_{S}$ is also multivariate Gaussian such that $ p(\mathbf{X}_{\bar{S}}| \mathbf{X}_{S} = \mathbf{x}_{s}) = \mathcal{N}_{\bar{S}}(\pmb{\mu}_{\bar{S}|S},\mathbf{\Sigma}_{\bar{S}|S})$ where $\pmb{\mu}_{\bar{S}|S} = \pmb{\mu}_{\bar{S}} + \mathbf{\Sigma}_{\bar{S}S} \mathbf{\Sigma}^{-1}_{SS} (\mathbf{x}_{S} - \pmb{\mu}_{S})$ and $\mathbf{\Sigma}_{\bar{S}|S} = \mathbf{\Sigma}_{\bar{S}\bar{S}} + \mathbf{\Sigma}_{\bar{S}S} \mathbf{\Sigma}^{-1}_{SS} \mathbf{\Sigma}_{S\bar{S}}$. We can therefore sample from the conditional Gaussian distribution with expectation vector and covariance matrix given by $\pmb{\mu}_{\bar{S}|S}$ and $\mathbf{\Sigma}_{\bar{S}|S}$ where $\pmb{\mu}$ and $\mathbf{\Sigma}$ are estimated by the sample mean and covariance matrix of $\mathbf{X}_{input}$.

\subsection{Synthetic Experiment: Interaction in the Model}

\begin{table}[t]
\label{exp_model_int}
\centering
\caption{Mean Average Error $\pm$ std, for SS and SV attributions under $v_{marg}$ for the three functions outlined in Section 5.1. SS perfectly identifies NSVGs for all three functions. \newline}
\begin{tabular}{|l|l|l|ll}
\cline{1-3}
        & SS & Shapley Value   &  &  \\ \cline{1-3}
$f_{1}$ & $\mathbf{0.000 \pm 0.000}$    & $0.335 \pm 0.400 $ &  &  \\ \cline{1-3}
$f_{2}$ & $\mathbf{0.000 \pm 0.000}$    & $1.143 \pm 0.990$ &  &  \\ \cline{1-3}
$f_{3}$ & $\mathbf{0.000 \pm 0.000}$    & $0.540 \pm 0.580$ &  &  \\ \cline{1-3}
\end{tabular}
\end{table}

We first construct three functions with linear and non-linear feature interactions:
$$f_{1}(\mathbf{X}) = X_{0} + (X_{1} / (2 + X_{4})) + 2(X_{2} * X_{3}) + sin(2(X_{5}) + X_{6})$$
$$f_{2}(\mathbf{X}) = 2(sgn(X_{0})) + sgn(X_{1} X_{2} X_{3}) + sgn(X_{4} X_{5} X_{6})$$
$$f_{3}(\mathbf{X}) = 2(X_{0}X_{2}X_{3}) + 4(X_{4} X_{5}) - 3(X_{1})^2 - (X_{6}))$$

We construct a synthetic dataset of seven features drawn independently from $\mathcal{N}(-1,1)$. For each of 100 randomly drawn samples we compute SS under $v_{marg}$. As $|\mathbf{X}| = 7$ we are able to compute the true SVs under $v_{marg}$ for each feature, without relying on a sampling algorithm. As we know the ground truth we calculate the Mean Average Error across all features and samples as our evaluation metric, 

\begin{equation} \label{eq_MAE}
MAE = \frac{1}{k} \sum^{k}_{j=1} \frac{1}{n} \sum_{i=1}^{n} m(X_{ij}) - gt(X_{ij}),
\end{equation}

where $m(X_{ij})$ is the attribution given by $m=SS$ or $m=SV$ to feature $i$ in sample $k$. As SS calculates an attribution for a set of features, $m_{SS}(X_{ij}) = \varphi_{\mathbf{X_{ij}}}$, the ground truth attribution $gt(X_{ij})$ is the ground truth value of each NSVG. For example, given $f= 2(X_{1}X_{2})$ and $\mathbf{x}_{j} = (1,1)$, $gt(X_{1,j}) = 2$ and $gt(X_{2,j}) = 2$.

Results are shown in Table 1. SS is successful in decomposing each function into its NSVGs and the attributions awarded to each set matches the ground truth of the function giving MAE of zero for all samples and functions. SV attributions deviate from ground truth by dividing the value of each NSVG between each individual feature which results in misleading attributions, particularly in the presence of inverse relationships between features. For example consider the following sub-component $(X_{1})/(1 - X_{2})$, and a particular sample $\mathbf{x}=(1,0.2)$. SV gives $X_{1}$ a positive attribution but $X_{2}$'s attribution is negative. Under SS, $X_{1}$ and $X_{2}$ are considered as non-separable and awarded a positive attribution together. From its SV attribution, a user may opt to change $X_{2}$ rather than $X_{1}$, however, as these features jointly move the outcome from the baseline to the target, the impact of changing $X_{2}$ in isolation could be cancelled out by the impact of $X_{1}$.

\subsection{Synthetic Experiment: Interaction in the Data}

\begin{table}[t]
\centering
\label{table_model_int}
\caption{Mean Average Error $\pm$ std for SS under $v_{cond}$ and SV under $v_{cond}$ and $v_{marg}$ for the three experiments outlined in Section 5.2. SS has lower MAE than SV for all models \newline}
\begin{tabular}{|l|l|l|l|l}
\cline{1-4}
        & SS                                 & Shap Marg & Shap Cond &  \\ \cline{1-4}
$g_{1}$ & $\mathbf{0.204 \pm 0.114}$                   & $0.226 \pm 0.121$        & $ 0.211 \pm 0.127$          &  \\ \cline{1-4}
$g_{2}$ & $\mathbf{0.071 \pm 0.031}$ & $0.082 \pm 0.032$        & $0.073 \pm 0.031$           &  \\ \cline{1-4}
$g_{3}$ & $\mathbf{0.074 \pm 0.044}$                   & $0.110 \pm 0.068$        & $ 0.150 \pm 0.059 $         &  \\ \cline{1-4}
\end{tabular}
\end{table}

We adopt the approach of \cite{hooker2021unrestricted} and propose and underlying linear regression model $f(\mathbf{X}) = X_{0} + 0.5 X_{1} + 0.8X_{3} + 0.2 X_{2} + 0.5 X_{4}$. We construct a synthetic dataset comprising five features $n=5$. $(X_{2},X_{3},X_{4})$ are all modeled as I.I.D and drawn independently from $\mathcal{N}(-1,1)$. $X_{0},X_{1}$, however are modeled as dependent features where $X_{1} = \rho X_{0}$. We generate a synthetic dataset $X_{train},X_{test}$ consisting $k=(2000,100)$ samples of each feature and obtain the ground truth labels $\mathbf{y}_{train},\mathbf{y}_{test} = f(\mathbf{X}_{train}),f(\mathbf{X}_{test})$.
We next select a model $g$ which is trained on $\mathbf{X}_{train},\mathbf{y}_{train}$ to approximate $f$. We calculate the attributions for each sample in $\mathbf{X}_{test}$ generated by the SV under both $v_{marg}$ and $v_{cond}$ and the attributions from SS under $v_{cond}$. To evaluate attributions we use the coefficients of the linear regression model as our ground truth attributions $c = \{1,0.5,0.8,0.2,0.5\}$. We use $MAE$ (Equation \ref{eq_MAE}) where the ground truth for feature $i$ in sample $j$ $gt_{X_{ij}} = c_{i} x_{i,j}$. 

Off-manifold attributions in the presence of interaction in the data recover the ground truth attributions reliably when $g$ is a linear model, however, that breaks down when non-linear models are used as the approximating function $g$ \cite{hooker2021unrestricted}. We therefore compare attributions under $g1$, a linear regression model, and $g2$, an XGBoost model.

Results are shown in Table 2 where SS outperforms SV on both $g1$ and $g2$. Under $g1$, the MAE is lower for SV Marginal than for SV Conditional, validating the findings in \cite{hooker2021unrestricted}. 

However, when non linear $g2$ is used, the attributions from SS and SV under $v_{cond}$ outperform SV under $v_{marg}$. The attributions provided by SS outperform those generated by SV across both models. We now show experimentally the claim that SS under on-manifold value function avoid the issues related to sensitivity. To do this we add a dummy variable $X_{5} = X_{0}$ to the dataset $\mathbf{X}$ such that $X_{5}$ is not used by $f$. We train another XGBoost model, $g3$ using the new dataset and generate the three sets of attributions as before. Results are shown in Table 2. Under the influence of the dummy, MAE of SV under $v_{cond}$ increases, as the attribution of each of the non-dummy variables moves further away from its true value to accommodate the attribution of the new feature despite it having no effect on the true output. In contrast, as SS includes this dummy feature in the non-separable set $\{X_{0},X_{1}\}$. The resulting attribution to the existing features is unchanged and thus the MAE remains constant under the inclusion dummy variables, demonstrating SS's robustness to how the underlying phenomenon is modelled. 

\begin{table}[t]
\centering
\caption{Average deletion $\pm$ std for the attributions generated by SS under $v_{marg}$ and $v_{cond}$, KS and TS for the Boston (B), Diabates (D) and Correlation (C) datasets. SS attributions have lowest deletion score across all datasets. \newline}

\begin{tabular}{|l|l|l|l|l|}
\hline
           & SS Int                     & SS Cond                    & KS                & TS                 \\ \hline
B     & $0.020 \pm 0.022$          & $\mathbf{0.007 \pm 0.006}$ & $0.046 \pm 0.047$ & $0.047 \pm 0.048$  \\ \hline
D  & $0.081 \pm 0.075$          & $\mathbf{0.050 \pm 0.039}$ & $0.103\pm 0.085$  & $0.010 \pm 0.082$ \\ \hline
C & $\mathbf{0.005} \pm 0.007$ & $0.033\pm 0.029$           & $0.075\pm 0.057$  & $0.072 \pm 0.055$  \\ \hline
\end{tabular}
\label{table_deletion}
\end{table}

\subsection{Shapley Sets of Real World Benchmarks}
We now evaluate SS on real data: the Diabetes, Boston and Correlation datasets from the Shap library \cite{NIPS2017_7062}. For each dataset we train either an XGBoost or Random Forest model on the provided train set obtaining $R^{2}$ score of 0.90 (RF), 0.89 9 (RF) and 0.86 (XGB) respectively. We compute SS attributions for 100 randomly selected samples from the test set under both $v_{marg}$ and $v_{cond}$. As the dimensionality of the datasets now exceed that capable of being computed by the true Shapley values we compare the SS attributions with the most commonly used approximation techniques: Tree Shap (TS) \cite{lundberg2020local2global} and Kernel Shap (KS) \cite{NIPS2017_7062}. Under its original implementation, KS is an approximation of an off-manifold value function and breaks the relationship between input features and the data distribution. TS does not make this assumption and is presented as an on-manifold Shapley value approximation. However, in practice TS performs poorly when there is high dependence between features in the dataset \cite{aas2021explaining}. 
To evaluate the attributions generated by SS, KS and TS in the absence of a ground truth attribution we use modified versions of the deletion and sensitivity measures which have been used widely across the literature \cite{gevaert2022evaluating}. Deletion is built on the intuition that the magnitude of a feature’s score should reflect its impact on the output. Our metric therefore measures the absolute distance between the target prediction, 

$v(\mathbf{x},\mathbf{X}_{\varnothing})$ and the prediction of a given sample $v(\mathbf{x},\mathbf{X})$ after the most important feature $X'_{i}=x_{i}$, determined by the attribution method under consideration $m$, has been removed. 
\begin{equation}
AD = \frac{1}{k} \sum_{j=1}^{k} |v(\mathbf{x}_{j},\varnothing) - v(\mathbf{x}_{j},N \backslash \{i\})|
\end{equation}

\begin{table}[t]
\label{table_sensitivity}
\centering
\caption{Average sensitivity $\pm$ std for SS under $v_{marg}$ and $v_{cond}$, KS and TS for the Boston (B), Diabates (D) and Correlation (C) datasets. SS results in the lowest sensitivity for B and C yet KS achieves lowest sensitivity for D. \newline}
\begin{tabular}{|l|l|l|l|l|}
\hline
           & SS Int                     & SS Cond                    & KS                                         & TS                 \\ \hline
B     & $0.015 \pm 0.049$          & $\mathbf{0.006 \pm 0.031}$ & $0.029 \pm 0.000$         & $0.030 \pm 0.000$  \\ \hline
D   & $0.021\pm 0.099$           & $0.017 \pm 0.067$          & $\mathbf{0.004 \pm 0.000}$ & $0.076 \pm 0.000$ \\ \hline
C & $\mathbf{0.000 \pm 0.010}$ & $0.008\pm 0.020$           & $0.001 \pm 0.000$                         & $0.035\pm 0.000$  \\ \hline
\end{tabular}
\end{table}

Low AD indicates that the attribution technique has correctly identified an important feature to remove. As SS attributes to sets of features we allow $\mathbf{X}'$ to be a non-separable variable set as generated by SS. This may influence the reliability of AD due to a varying number of features being removed from an instance. We therefore also assess the sensitivity of the attribution technique which calculates the difference between the sum of all the attributions given by the attribution technique and the prediction of the sample. Ideal attributions have a low sensitivity. 
\begin{equation}
AS = \frac{1}{k} \sum_{j=1}^{k} |v(\mathbf{x}_{j},N) - \sum_{i = 1}^{n} m(\textbf{X}_{ij})|
\end{equation}
Tables 3 and 4 show how SS has lower (better) deletion than TS and KS across all three datasets. However, KS has the lowest sensitivity score on the Diabetes dataset, we note that for this dataset, there is high variance of the sensitivity score for both SS attributions. This can be largely explained by the sensitivity of SS to the setting of $\epsilon$ which is discussed further in Section 7. 
\begin{figure}[h]
\centering
\begin{minipage}{.5\textwidth}
  \centering
  \includegraphics[width=\linewidth]{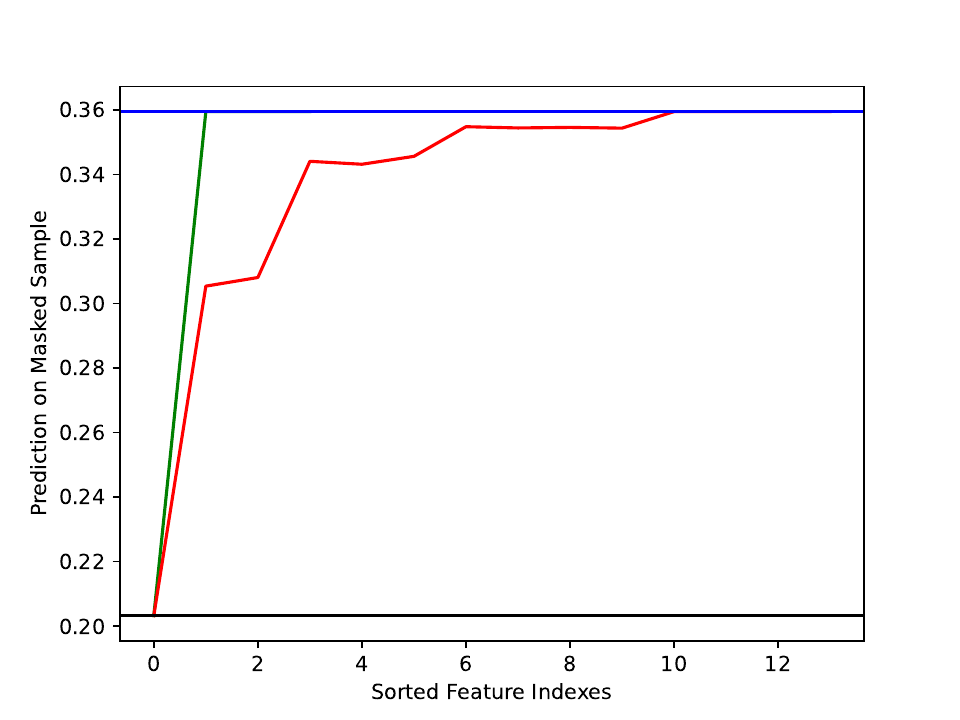}
  \label{fig:test1}
\end{minipage}%
\begin{minipage}{.5\textwidth}
  \centering
  \includegraphics[width=\linewidth]{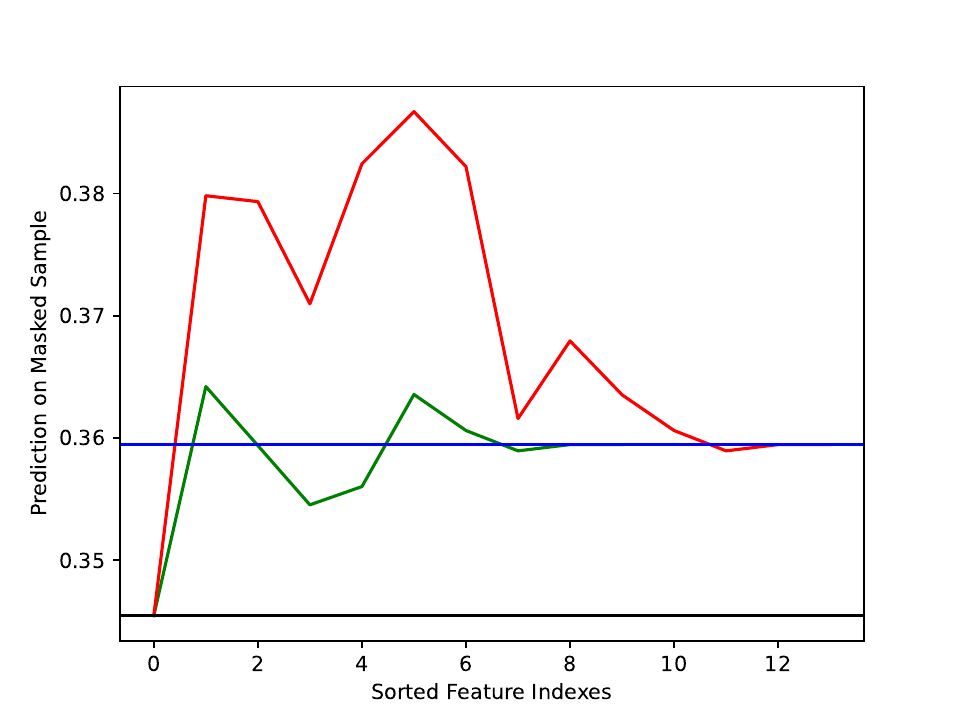}
  \label{fig:test2}
\end{minipage}
\caption{Curves show the change in prediction of two individual samples from the Boston dataset as increasing features, as sorted in order of importance by the attributions returned by SS (green) and KS (red) attributions are removed from the instance. Original and target predictions are shown by the black and blue horizontal line. An ideal attribution would result in a sharp increase or decrease towards the target. In both samples, SS results in a quicker and smoother transition from original to target prediction }
\label{fig_boston}
\end{figure}
Figure \ref{fig_boston} shows the advantage of sets rather than individual attributions. The red and green curves (KS and SS respectively) show the change in prediction as each feature in the sorted attributions is masked consecutively from the input.
By considering the effect of sets of interacting features rather than individual features we can see that SS avoids the sub-optimal behaviour of KS which arises due to the interaction effects between features in the model masking each other's importance. Figure \ref{fig_boston} also validates the use of the deletion to compare individual and set attributions as it is clear that masking more features does not guarantee a lower deletion score.

\section{Conclusions, Limitations and Future Work}
This paper has introduced Shapley Sets (SS), a novel method for feature attribution, which automatically and optimally decomposes a function $f$ into a set of NSVGs by which to award attribution. We have shown how SS generates more faithful explanations in the presence of feature interaction both in the data and in the model than Shapley value-based alternatives. To our knowledge, SS is the only method in the literature which automatically generates a grouped attribution vector. Below we explore some limitations of SS and ideas for future work. \textbf{Sensitivity to Parametrisation}: In Algorithm 2, $\epsilon$ determines the degree to which two sets of variables are considered interacting. The original RDG algorithms recommends the setting $\epsilon$ as proportional the magnitude of the objective space. This setting works well for SS Interventional. However, we noticed a large variation in the variable grouping generated by SS Conditional under this setting of $\epsilon$. This is not surprising as it is known that $v_{cond}$ is sensitive to feature correlations in the data and it is difficult to know how much correlational structure to allow before two features are considered to be causally linked. Future work should therefore look at alternative methods of function decomposition which are not so dependent on the parametrisation of $\epsilon$ \cite{chen2022decomposition}. \textbf{Assumption of Partially-Separable Model} SS assumes that the model to be explained is partially  separable. If we consider the function $f(\mathbf{X}) = X_{1} X_{2}X_{3}$, SS would result in a single attribution to all three features of $f(\mathbf{x})$. This is not useful from an explanation perspective although does inform us about the nature of the underlying model. Furthermore, the assumption of a partially separable function is also made by the Shapley value \cite{kumar2020problems}.  Future work should consider function decomposition under a wider class of separability such as multiplicative separability where associated algorithms decompose a function into its additive and multiplicative separable variable sets \cite{chen2022decomposition}. 

\newpage

\section*{Acknowledgments}
We would like to thank Giulia Occhini, Alexis Monks, Isobel Shaw, Jennifer Yates, for their invaluable support during the writing of this paper. This work was supported by an Alan Turing Institute PhD Studentship funded under EPSRC grant EP/N510129/1.

\bibliographystyle{unsrt}  
\bibliography{references}

\end{document}